\documentclass[journal, 10 pt,onecolumn, draftclsnofoot,]{IEEEtran}
\usepackage{graphicx}
\usepackage{cite}
\usepackage{picinpar}
\usepackage{amsmath}
\usepackage{ragged2e}
\usepackage{amsfonts}
\usepackage{url}
\usepackage{flushend}
\usepackage[latin1]{inputenc}
\usepackage{colortbl}
\usepackage{soul}
\usepackage{pifont}
\usepackage{color}
\usepackage{alltt}
\usepackage[hidelinks]{hyperref}
\usepackage{enumerate}
\usepackage{subcaption}
\usepackage{multirow}
\usepackage{siunitx}
\usepackage{breakurl}
\usepackage{epstopdf}
\usepackage{pbox}
\usepackage{booktabs}
\usepackage{mdwtab}
\usepackage{booktabs}
\usepackage[normalem]{ulem}
\usepackage{amsmath,amsfonts,amsthm}
\useunder{\uline}{\ul}{}

\newcommand\blfootnote[1]{%
  \begingroup
  \renewcommand\thefootnote{}\footnote{#1}%
  \addtocounter{footnote}{-1}%
  \endgroup
}
\begin{document}

\title{Attention Sequence to Sequence Model for Machine Remaining Useful Life Prediction}

\author{Mohamed~Ragab, Zhenghua~Chen, Min Wu, Chee-Keong Kwoh, Ruqiang Yan, and Xiaoli Li 
}

\markboth{Journal of \LaTeX\ Class Files,~Vol.~14, No.~8, August~2015}%
{Shell 
\MakeLowercase{\textit{et al.}}: Bare Demo of IEEEtran.cls for IEEE Transactions on Industrial Informatics Journals}
\maketitle
%
\blfootnote{
\indent Submitted to XXX}

\begin{abstract}
Accurate estimation of remaining useful life (RUL) of industrial equipment can enable advanced maintenance schedules, increase equipment availability and reduce operational costs. However, existing deep learning methods for RUL prediction are not completely successful due to the following two reasons. First, relying on a single objective function to estimate the RUL will limit the learned representations and thus affect the prediction accuracy. Second, while longer sequences are more informative for modelling the sensor dynamics of equipment, existing methods are less effective to deal with very long sequences, as they mainly focus on the latest information. To address these two problems, we develop a novel attention-based sequence to sequence with auxiliary task (ATS2S) model. In particular, our model jointly optimizes both
\textit{reconstruction loss} to empower our model with predictive capabilities (by predicting  next  input  sequence  given   current  input  sequence) and  \textit{RUL prediction loss} to minimize the difference between the predicted RUL and actual RUL. Furthermore, to better handle longer sequence, we employ the attention mechanism to focus on all the important input information  during training process. Finally, we propose a new \textit{dual-latent feature representation} to integrate the encoder features and decoder hidden states, to capture rich  semantic information in data. We conduct extensive experiments on four real datasets to evaluate the efficacy of the proposed method. Experimental results show that our proposed method can achieve superior performance over 13 state-of-the-art methods consistently. 
\end{abstract}

\IEEEpeerreviewmaketitle
\section{Introduction}

Prognostic and Health Management (PHM) is receiving much attention in many industrial applications, as it can  potentially reduce equipment downtime and increase system reliability. Typically, PHM systems are leveraged to monitor the condition of mechanical or electrical equipment based on their environmental information and domain knowledge. 

One key task in PHM is the reliable prediction of \textit{remaining useful life} (RUL) of an equipment. With accurate RUL estimation, industries can have predictive maintenance planning and thus prevent catastrophic failures or faults from happening \cite{sikorska2011prognostic}. Approaches for RUL prediction can be classified into two broad categories, namely, model-driven approaches~\cite{pecht2009physics,tamssaouet2019system},  data-driven approaches~\cite{si2011remaining} , and hybrid approaches~\cite{aizpurua2017model, daroogheh2016prognosis}.
Specifically, model-driven approaches require strong theoretical understanding to model the behaviour of equipment 
and its detailed degradation process. As equipment performance complexity continues to evolve, it becomes extremely challenging to apply  model-driven approaches in real applications. 

On the other hand, with increasing data availability in smart manufacturing, data-driven approaches have emerged more promisingly for predicting the RUL of equipment. Traditional machine learning models have been used to estimate the RUL, including hidden Markov model, artificial neural network \cite{ali2015accurate}, extreme learning machines \cite{javed2015new}, and support vector machines \cite{svr}. However, these approaches suffer from the extensive efforts for feature engineering. Deep learning with the ability of automatic feature extraction has achieved wide success in many fields, including computer vision, natural language processing, and speech recognition \cite{bengio2017deep}. Very recently, various deep learning methods, e.g., convolutional neural network (CNN) and recurrent neural network (RNN), have also been explored for RUL prediction \cite{zhao2019deep}. 
 
 
For instance, Li \textit{et al.,} proposed a CNN with 1-D filters to extract features from input sensor data for RUL prediction and also used window-time approach to prepare data samples for improved feature extraction~\cite{dcnn}. Yang \textit{et al.,} developed a two-stage approach by using one CNN network to inspect the fault points and another CNN to estimate the RUL \cite{yang2019remaining}. Zhu \textit{et al.,} proposed a multi-scale CNN to extract features and predict the degradation of bearings~\cite{zhu2018estimation}. Zhang \textit{et al.,} combined multi-layer perception (MLP) and CNN to extract features from vibration data and predict the health index of machines~\cite{zhang2019roller}. As shown in above studies, CNN based methods have achieved good performance for RUL prediction. However, they have limitations when dealing with the sequence data as they ignore the temporal dependency among data points in a given input sequence. 

Recurrent neural networks (RNN) have been shown to be effective in modeling dynamic systems and learning temporal dependency in data. In particular, Long Short-Term Memory (LSTM) is a special type of recurrent model that can model the dynamics of sequences by introducing the memory cells \cite{hochreiter1997long}. It has become increasingly popular for RUL prediction. For instance, Zheng \textit{et al.,} have used two layers LSTM network to predict the RUL of turbofan engines \cite{lstm}. Huang \textit{et al.,} employed a stacked-bidirectional LSTM with auxiliary inputs to model sensor data under multiple operating conditions \cite{Bilstm}. Miao \textit{et al.,} designed a deep LSTM framework to jointly perform degradation assessment and RUL prediction \cite{miao2019joint}.

Other recent approaches have combined the LSTM networks with CNN networks for RUL prediction. For example, Al-Dulaimi \textit{et al.,} proposed a two-parallel path approach with one for LSTM and one for CNN \cite{hdnn}. Liu \textit{et al.,} combined CNN with LSTM in a series manner and fed the output convolutional features to a bi-directional LSTM network in order to improve the latent representation of the input sequence~\cite{liu2019a}.
 
In addition to CNN and LSTM based methods, other deep learning algorithms have also been developed for RUL prediction. Min \textit{et al.,} presented denoising autoencoder based deep neural networks (DNNs) with a two-stage approach to estimate the RUL of bearings~\cite{xia2019a}. Ma \textit{et al.,} used coupling autoencoder model on multimodal sensor data to perform fault diagnosis~\cite{ma2018deep}. In addition, a deep belief network (DBN) is proposed together with ensemble techniques for RUL prediction \cite{dbn}. Deutsch \textit{et al.,} integrated a deep belief network with a fully connected network to predict the RUL for rotating components \cite{deutsch2017using}. Liao \textit{et al.,}, employed restricted Boltzmann machine (RBM) to automatically extract features for RUL prediction \cite{rbm}. Encoder-decoder networks (e.g., LSTM-ED \cite{lstm-ed} and BiLSTM-ED \cite{bilstm-ed}) have also been employed for health index prediction and RUL estimation.

 
Although these methods showed great potential for RUL estimation problem, 
there still some shortcomings to be addressed:
 \begin{itemize}
 \item LSTM tends to lose relevant and important historical information 
 when dealing with very long sequences~\cite{cho2014on}. It focuses on those latest sequence information when mapping the whole input sequence into fixed-length vector representation. 
 \item Many related deep learning approaches rely only on single objective, i.e., the regression objective on RUL label, to extract the features and predict the RUL. We argue that the representation learning  can be improved by being less focused on single supervised objective \cite{trinh2018learning}.
 
 \end{itemize} 
 
To address the above two problems, we propose a dual-objective sequence to sequence approach named ATS2S for accurate RUL prediction. We employ the sequence to sequence based learning model for two objectives concurrently: (1) reconstruct the \textit{next} input sequence from given input sequence; (2) predict the RUL of the given input sequence. 
In particular, the sequence to sequence model aims to reduce sensor noise by compressing the information from the input sequence into a fixed-length vector. Note that it is challenging for the network to handle very long sequences, as the prediction performance may deteriorate rapidly with the increase of the input sequence length~\cite{cho2014on}. To tackle this issue, we propose an attention based decoding and focus on the important parts of the input sequence (instead of the latest information in LSTM) that can maximize the decoding performance without losing relevant information. In addition, we integrate the last hidden state of the decoder with the encoder hidden features, as a comprehensive \textit{dual-latent feature representation} for the RUL predictor. Note that there are encoder-decoder based approaches (e.g., LSTM-ED \cite{lstm-ed} and BiLSTM-ED \cite{bilstm-ed}) for RUL prediction in the literature. Our proposed ATS2S is different from them in the following aspects: First, ATS2S is an end-to-end framework, while their methods extract features and predict RUL separately. Second, ATS2S implements an attention mechanism and leverages the \textit{dual-latent feature representation} for RUL prediction, while their methods still use the encoder's last hidden state as features for health index prediction and RUL estimation.

Overall, our main contributions can be summarized as follows.

 \begin {itemize}

 \item Our model jointly optimizes both
\textit{reconstruction loss} to empower our model with predictive capabilities (by predicting the next  input  sequence  given   current  input  sequence) and  \textit{RUL prediction loss} to minimize the difference between the predicted RUL and actual RUL. 
 \item We design an \textit{attention mechanism} in the encoder-decoder network to handle the long sequences. As such, our model can focus on the most relevant information of the input sequences for RUL prediction.

\item We propose a new \textit{dual-latent feature representation} to integrate the encoder features and decoder hidden states, to capture rich  semantic information in data.  

\item We conduct extensive experiments on four benchmark datasets to evaluate our proposed approach. The results show that the proposed approach can significantly improve RUL prediction over 14 state-of-the-arts. 
 \end{itemize}


\begin{figure*}[ht!]
\centering
 \includegraphics[width=1\textwidth]{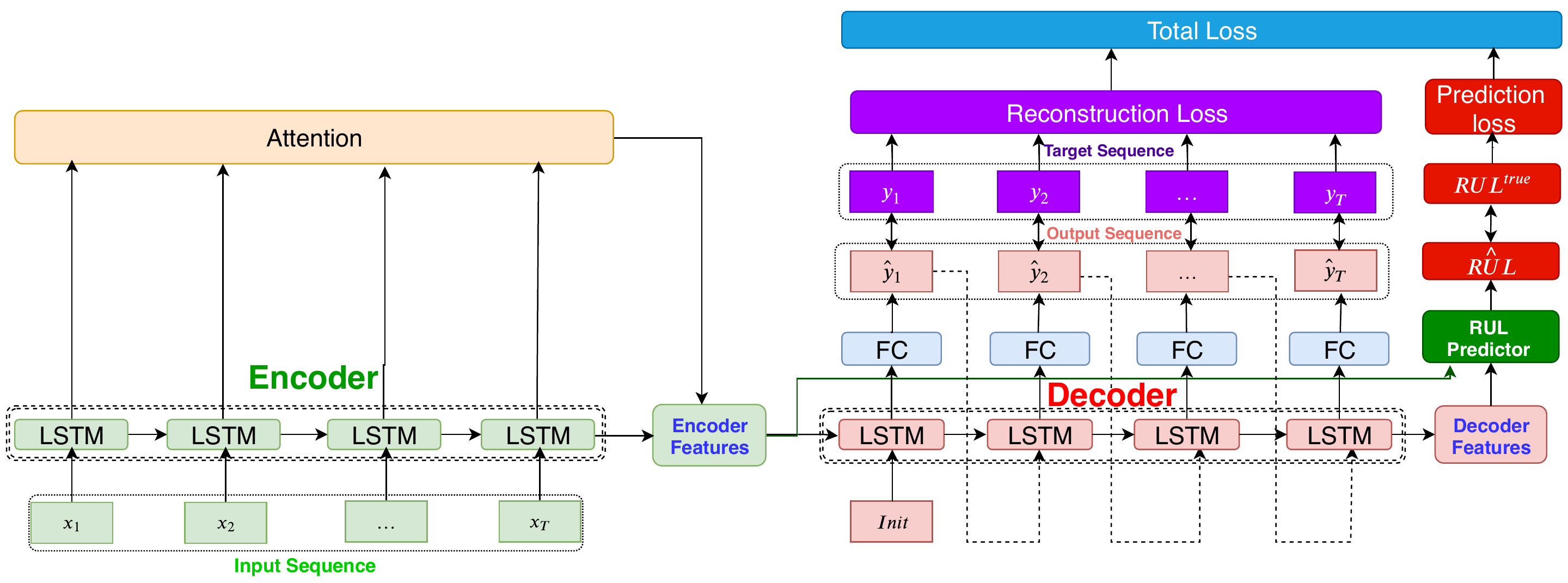}
  \caption{Attention-based sequence to sequence model for RUL prediction}
  \label{fig:DTS2S}
\end{figure*}

\section{Methodology}

In this section, we will introduce our proposed attention-based sequence to sequence with auxiliary task (ATS2S) model for RUL prediction.

\subsection{Overview of ATS2S}

The proposed ATS2S is composed of three main components, namely, encoder, decoder, and RUL predictor, as shown in Fig.~\ref{fig:DTS2S}. 
Firstly, the encoder maps the whole input sequence into a sequence of hidden states. Unlike conventional encoder-decoder models that compress all the input information into the single fixed-length vector (i.e., encoder's last hidden state), we design an \textit{attention layer} to select the hidden states that are relevant and important for the decoding (removing noise). Then, we pass the weighted sum of the encoder hidden states (i.e., attention outputs) as \textit{encoder features} to decoder. The decoder is then trained to forecast the \textit{next} input sequence given the current input sequence, in order to give our model more \textit{predictive power}. 
Finally, the RUL prediction network (a fully connected neural network) takes \textit{dual-latent feature representation} to integrate both the encoder and decoder hidden states/features for RUL prediction. The predictor maps from the feature dimension space to a single value, i.e., predicted RUL.

Note that our ATS2S method jointly optimizes the RUL prediction loss, which is the difference between the predicted RUL label and ground-truth label, as well as the reconstruction loss, which is the difference between predicted and actual sequence.  
Next, we will introduce each of the three components of ATS2S in details.

\subsection{LSTM Based Encoder}
In order to model the input dynamics of sensor signals, we employ the LSTM model as our backbone architecture in the sequence to sequence model. Given an input sample $\textbf{X} =(\textbf{x}_1, \textbf{x}_2,\dots,\textbf{x}_T) \in {\mathbb{R}}^{n\times T}$, $\textbf{x}_t \in {\mathbb{R}}^n$ is n-dimensional input vector at each time step $t$ ($1 \leq t \leq T$) from $n$ sensors. At each time step $t$, LSTM takes the input vector $\textbf{x}_t$ and \textit{previous} hidden state $\textbf{h}_{t-1}$ to produce \textit{current} hidden state $\textbf{h}_t$, current long term memory cell $\textbf{c}_t$ and output $\textbf{o}_t$. The following equations demonstrate the detailed process in the LSTM cell. 
\begin{align}
\textbf{i}_t =\sigma(\textbf{W}_i\textbf{x}_t+\textbf{U}_i\textbf{h}_{t-1}+\textbf{b}_i ), \\
\textbf{f}_t= \sigma(\textbf{W}_\textbf{f}\textbf{x}_t+\textbf{U}_f\textbf{h}_t-1+\textbf{b}_f ),  \\
\textbf{o}_t= \sigma(\textbf{W}_o\textbf{x}_t+\textbf{U}_o\textbf{h}_t-1+\textbf{b}_o ),  \\
\textbf{g}_t=  \textit{tanh}(\textbf{W}_c\textbf{x}_t+\textbf{U}_c\textbf{h}_t-1+\textbf{b}_c ),\\ 
\textbf{c}_t=\textbf{f}_t\odot \textbf{c}_t-1 +\textbf{i}_t\odot \textbf{g}_t, \\
\textbf{h}_t= \textbf{o}_t \odot tanh(\textbf{c}_t), 
\end{align}
where $\sigma$ is nonlinear \textit{sigmoid} function, $\odot$ is an element-wise multiplication operator, $\textbf{W}_*\in{\mathbb{R}}^{n\times p}$ (i.e., $\textbf{W}_i$, $\textbf{W}_f$, $\textbf{W}_o$ and $\textbf{W}_c$) are the model parameters that map from input dimension $n$ to hidden dimension $p$, $\textbf{U}_*\in{\mathbb{R}}^{p\times p}$ map from the previous hidden dimension to the current hidden dimension, and $\textbf{b}_*\in{\mathbb{R}}^{p}$ are bias vectors. It worth noting that the parameters are shared across all the time steps. 
The Encoder model $f_{enc}$ takes the input sequence $(\textbf{x}_1,\textbf{x}_2,\dots,\textbf{x}_T)$ and produces a sequence of hidden states ($\textbf{h}_1,\textbf{h}_2,\dots,\textbf{h}_T$) and a sequence of cell states ($\textbf{c}_1,\textbf{c}_2,\dots,\textbf{c}_T$) in Equation (\ref{equ:encoder}). 

\begin{align}
[(\textbf{h}_1,\dots,\textbf{h}_T),(\textbf{c}_1,\dots,\textbf{c}_T)]= f_{enc}(\textbf{x}_1,\textbf{x}_2,\dots,\textbf{x}_T; \boldsymbol{\theta}_{enc}),
\label{equ:encoder}
\end{align}
where $\boldsymbol{\theta}_{enc}=[\textbf{W}_{enc},\textbf{U}_{enc},\textbf{b}_{enc}]$ are the parameters of the Encoder model. 

\subsection{Attention Based Decoding}

The main idea of attention is inspired by human visual systems where human can focus on the relevant part of a scene and ignore irrelevant parts. 
Similarly, we design an attention mechanism in our sequence to sequence model for the whole sequence of hidden states. In particular, we focus on \textit{all the important} hidden states of the encoder for decoding, while standard sequence to sequence model relies solely on the \textit{last} hidden state and thus loses valuable information. 

\begin{figure}[htbp]
    \centering
    \includegraphics[width=0.4\textwidth]{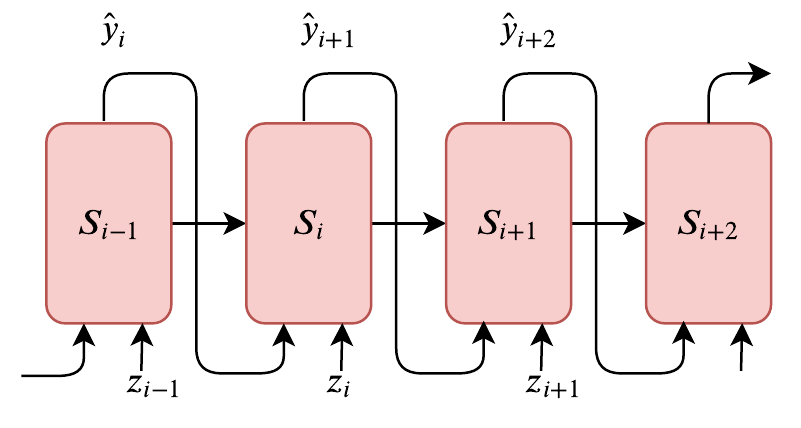}
    \caption{Decoding based on the encoder context vector at each time step}
    \label{decode}
\end{figure}

More specifically, at each time step $i$, the decoder model $f_{dec}$ takes three inputs, i.e., context vector $\textbf{z}_i$, \textit{previous} decoder hidden state $\textbf{s}_{i-1}$, and input $\hat{\textbf{y}}_i$, to produce the \textit{current} decoder hidden state $\textbf{s}_i$, as shown in Fig.~\ref{decode}. 
We calculate the decoder output according to the following equation:
\begin{align}
\textbf{s}_i=f_{dec}((\hat{\textbf{y}}_i,\textbf{z}_i,\textbf{s}_{i-1});\boldsymbol{\theta}_{dec}).
\end{align}
Then, we map from $\textbf{s}_i$ to the next step of the target $\hat{\textbf{y}}_{i+1} $ in Equation (\ref{equ:fc-de}):
\begin{align}
\hat{\textbf{y}}_{i+1}= f_p({\textbf{s}_i};\boldsymbol{\theta}_{l}),
\label{equ:fc-de}
\end{align}
where $f_p$ a function represents fully connected (FC) as shown in Fig. \ref{fig:DTS2S}, which maps from the hidden dimension to the output dimension. 

\begin{figure}[htbp]
    \centering
    \includegraphics[width=0.4\textwidth]{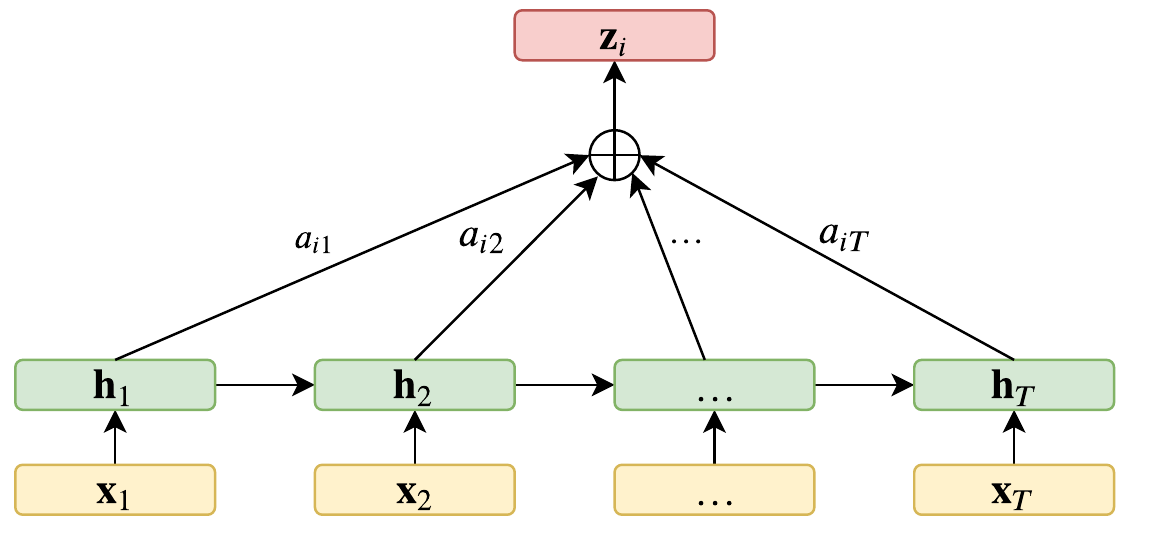}
    \caption{Applying attention on the encoder hidden states}
    \label{fig:attn_apply}
\end{figure}

The encoder features are defined as context vector $\textbf{z}_i$,  calculated as follows:
\begin{align}
\textbf{z}_i=\sum_{j=1}^{j=T} a_{ij}\textbf{h}_j,
\end{align}
where $\textbf{h}_j$ is the encoder's hidden state at position $j$, $a_{ij}$ is the attention weights that determine the importance of $\textbf{h}_j$ to $\textbf{z}_i$, and $\textbf{z}_i$ is the attention output, i.e., the weighted sum of the encoder's hidden states as shown in Fig.~\ref{fig:attn_apply}, which is able to capture all the relevant historical signals, instead of just focusing on the latest information used in LSTM. 
We  compute $a_{ij}$ as follows: 
\begin{align}
a_{ij} & =softmax({e}_{ij})=\frac{exp({e}_{ij})}{\sum_i exp({e}_{ij})},\\
{e}_{ij} & =f_{attn}((\textbf{s}_{i-1},\textbf{h}_j);\boldsymbol{\theta}_{attn}),
\end{align}
where $f_{attn}$ is a feed froward neural network that produces the alignment scores between $\textbf{h}_j$ and $\textbf{s}_{i-1}$.

\subsection{RUL Predictor}

\begin{figure}[htbp!]
    \centering
    \includegraphics[width=0.3\textwidth]{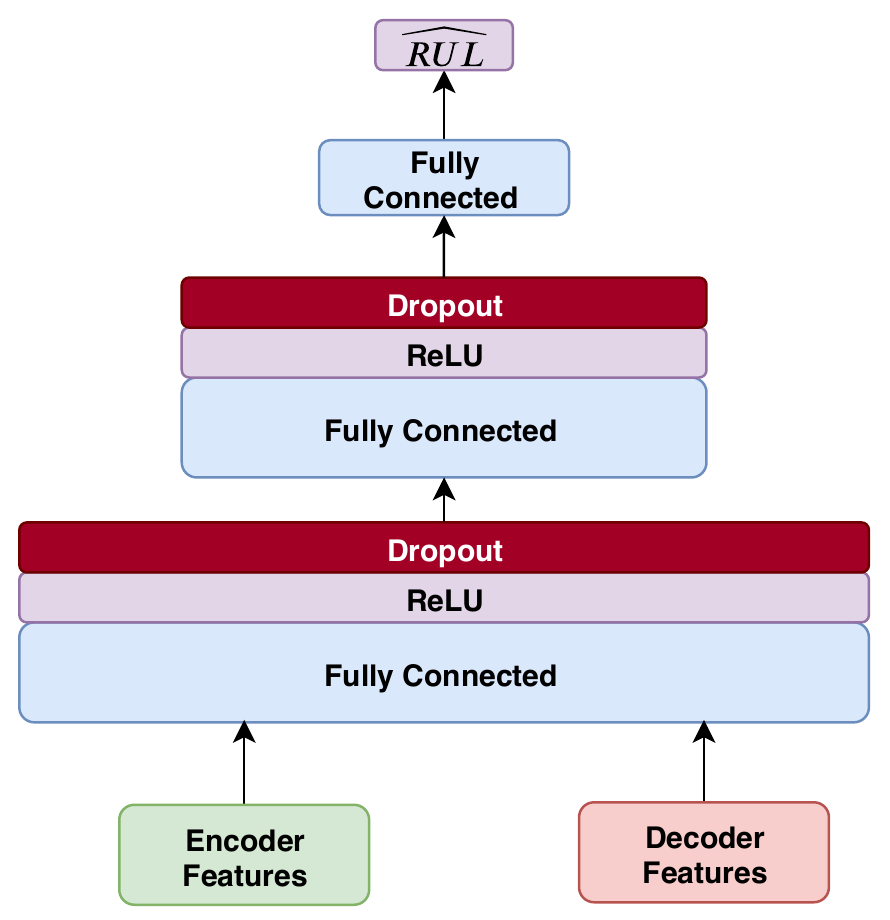}
    \caption{Architecture of RUL predictor network }
    \label{rul_pred}
\end{figure}

The objective of the RUL predictor is to accurately predict the corresponding RUL value for each input sequence (sensor signals). We first integrate the last hidden state of the decoder with the encoder hidden features, as a comprehensive \textit{dual-latent feature representation}, and then design
a function that maps the dual-latent feature representation to a single RUL value. We denote the RUL predictor as $f_{pred}:\mathbb{R}^D \rightarrow \mathbb{R}$ in Equation (\ref{equ:rul}), where $D$ is the dimension of dual-latent feature representation. 

\begin{align}
\widehat{RUL}= f_{pred}((\textbf{h}_T,\textbf{s}_T);\boldsymbol{\theta}_{pred}),
\label{equ:rul}
\end{align}
where $\widehat{RUL} \in \mathbb{R}$ is the predicted label, $\textbf{h}_T$ and $\textbf{s}_T$ are the features of encoder and decoder respectively. Fig. \ref{rul_pred} shows the diagram of the RUL predictor, which is a multi-layer feed-forward network followed by a non-linear activation function (i.e., ReLU). 

\subsection{Multi-objective  Optimization}

\subsubsection{Reconstruction Loss}
In our ATS2S, we aim to forecast the next input sequence given the current input sequence so that our model has predictive power. Therefore, we define the reconstruction loss as the mean square error between the target output and predicted output in Equation (\ref{equ:rec-loss}). In particular, 
$\textbf{Y}_{i} = 
(\textbf{y}_{1}, 
\textbf{y}_{2}, 
\cdots, 
\textbf{y}_{T}) \in \mathbb{R}^{n\times T}$, where $\textbf{y}_{t} = \textbf{x}_{t+1} \in \mathbb{R}^n$,  $1 \leq t \leq T$, $T$ is the length of the sequence, and $n$ is the number of sensors.
\begin{align}
L_{rec}(\theta)= \frac{1}{N}\sum_{i=1}^{N}||\hat{\textbf{Y}}_{i}-\textbf{Y}_{i}||_2^2,
\label{equ:rec-loss}
\end{align}
where $\textbf{Y}_i$ is target sequence, $\hat{\textbf{Y}}_i$ is the predicted sequence, $\theta$ is the model parameters, and $N$ is the total number of samples. 
\subsubsection{RUL Prediction Loss}
The RUL prediction loss is defined as the mean square error between the true RUL label and the predicted RUL label for each input sequence. The RUL loss can be defined as follows: 
\begin{align}
L_{rul}(\theta)= \frac{1}{N}\sum_{i=1}^{N}(\widehat{RUL}_i-RUL_i)^2
\end{align}
where $\widehat{RUL}_i$ is predicted label and $RUL_i$ is the true label.

\subsubsection{Joint Loss}

The proposed model aims to optimize both reconstruction and prediction losses concurrently. We argue that jointly optimizing both losses can not only provide a good and rich latent representation, but also improve the accuracy of RUL prediction. The joint loss can be formulated as follows

\begin{align}
L(\theta)= \alpha L_{rec}(\theta)+L_{rul}(\theta),
\label{joint_loss}
\end{align}
where $\alpha$ is a tunable parameter to control the contribution of the reconstruction loss. It can control the contribution from reconstruction loss while maintaining the prediction loss (the major loss for RUL prediction). 

\section{Experiments and Results}
We have conducted extensive experiments on benchmark data to evaluate the performance of our proposed model. 
\subsection{Experimental Data}

\begin{figure}[htbp!]
    \centering
    \includegraphics[width=0.4\textwidth]{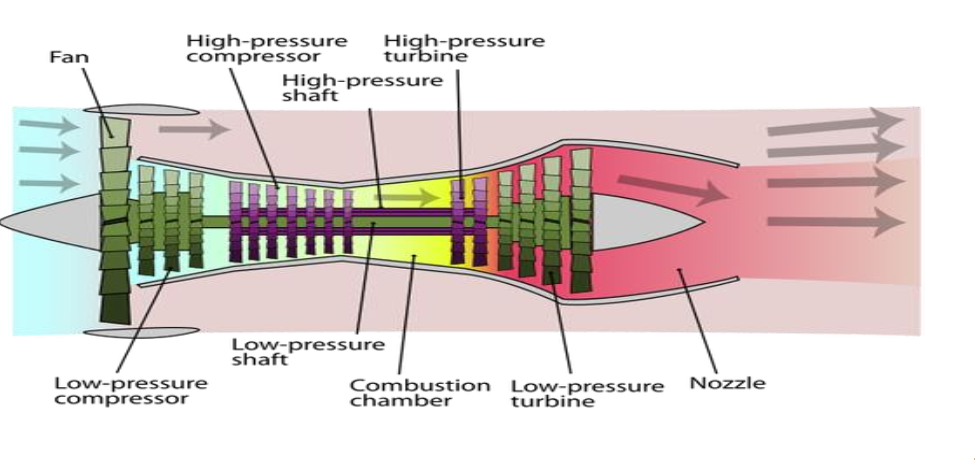}
    \caption{Diagram of the engines in C-MAPSS data \cite{cmapps}.}
    \label{fig:cmapps}
\end{figure}

We evaluate our proposed ATS2S method on C-MAPSS (Commercial Modular Aero-Propulsion System Simulation) data \cite{cmapps}. C-MAPSS data describes the degradation process of aircraft engines as shown in Fig. \ref{fig:cmapps}. It consists of four benchmark datasets with different number of training/testing engines, operating conditions and fault types. The details about these four datasets are summarized in Table~\ref{tab:cmapps_data}. 

\begin{table}[!h]
\centering
\caption{Properties of C-MAPSS Dataset }
\label{tab:cmapps_data}
\begin{tabular}{@{}lcccc@{}}
\toprule
Dataset & FD001 & FD002 & FD003 & FD004 \\ \midrule
\# Training engines & 100 & 260 & 100 & 249 \\ \midrule
\# Testing engines & 100 & 259 & 100 & 248 \\ \midrule
\# Operating conditions & 1 & 6 & 1 & 6 \\ \midrule
\# Fault types & 1 & 1 & 2 & 2 \\ \bottomrule
\end{tabular}%
\end{table}

\subsubsection{Sensor Data Selection} 
\begin{figure}[htbp]
    \centering
    \includegraphics[width=0.48\textwidth]{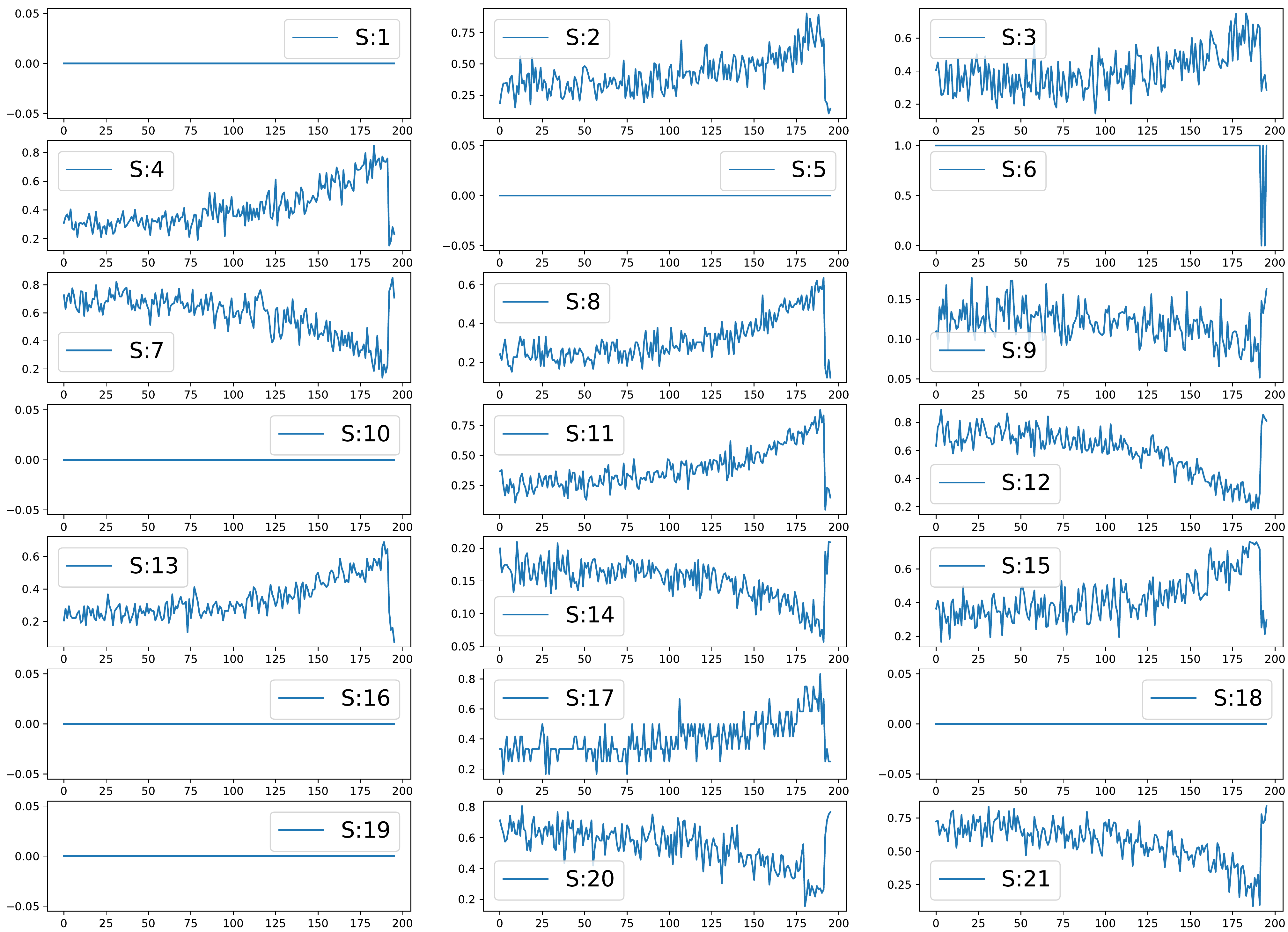}
    \caption{Degradation trend of one engine across 21 sensors on FD001.}
    \label{fig:sensors_reading_FD001}
\end{figure}

Twenty-one sensors are deployed in different locations of the engine to measure temperature, pressure and speed. To select relevant sensors for RUL prediction, we visualize the signals from all the 21 sensors for FD001.  Fig.~\ref{fig:sensors_reading_FD001} shows the sensor readings for a randomly selected engine. While most of sensors have a clear degradation trend, other sensors remain constant in the run-to-fail experiments (i.e., sensors 1, 5, 6, 10, 16, 18 and 19). Therefore, 14 sensors, namely sensors 2, 3, 4, 7, 8, 9, 11, 12, 13, 14, 15, 17, 20 and 21, are used for RUL prediction. FD003 follows the same degradation patterns as FD001 and thus we use the same subset of sensors for FD001 and FD003. Similar procedure has been done for FD002 and FD004. Eventually we adopt 9 sensors ~\cite{Bilstm}, namely sensors 3, 4, 9, 11, 14, 15, 17, 20 and 21, for RUL prediction on FD002 and FD004. 

\subsubsection{Data Segmentation and Processing} 
\begin{figure}[htbp]
    \centering
    \includegraphics[width=0.48\textwidth]{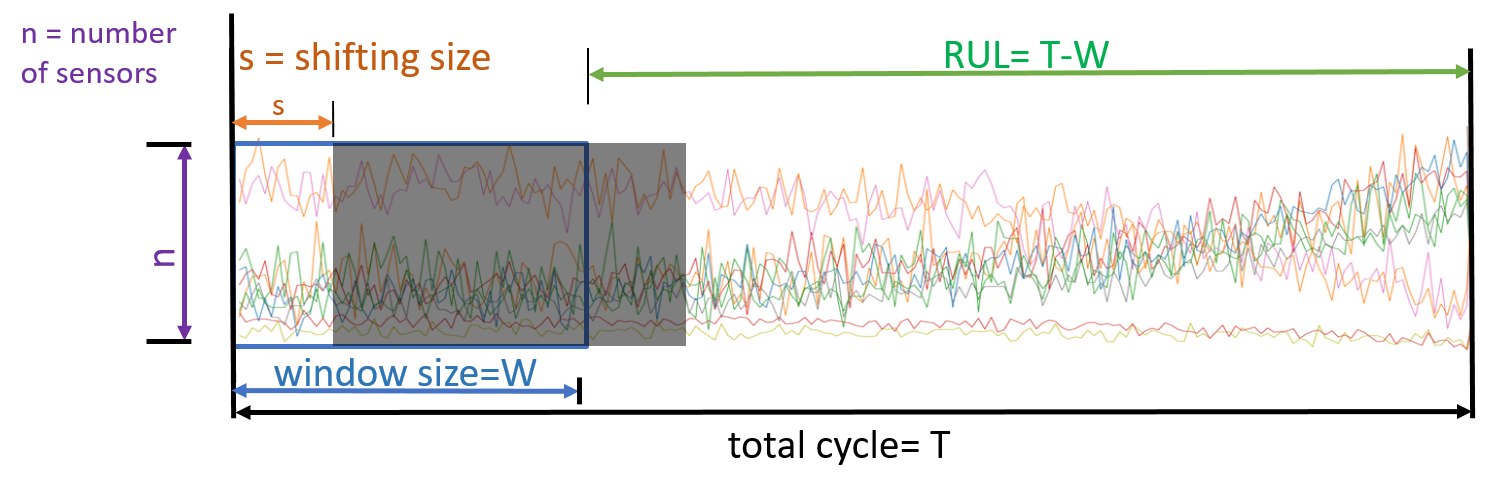}
    \caption{Data segmentation using sliding window for RUL prediction}
    \label{fig:moving_window}
\end{figure}

We follow the sliding window method ~\cite{babu2016deep, dbn} for data segmentation. Fig.~\ref{fig:moving_window} shows the process of data segmentation with sliding window, where $W$ is the window size, $n$ is the number of sensors and $s$ is the shifting size. Given that the total number of cycles is $T$, the RULs for the first and second windows/samples are thus $T-W$ and $T-W-s$, respectively. In our experiments, $W$ and $s$ are set to be 30 and 1, respectively.

Moreover, we adopt the piece-wise linear degradation model ~\cite{Bilstm,hdnn} for the RUL labels. In case a sample has a RUL value greater than a pre-defined threshold, we re-set the RUL value as the threshold for this sample. In particular, we follow the previous studies ~\cite{Bilstm,hdnn} and set the threshold as 125 for FD001/FD003 and 130 for FD002/FD004. 

\subsubsection{Data Normalization} The prognostic problem of real systems involves different types of sensors and different operating conditions. Directly feeding the raw sensor readings with high variance to the machine learning models may hinder the learning process and affect the model performance. To remedy this issue,  we use Min-Max normalization for each sensor restrict the values within $[0,1]$. For datasets with multiple working conditions, we normalize the sensor readings 
with respect to their corresponding working condition. In particular, we first group the sensors by their corresponding working conditions, then we apply normalization on each cluster independently. To formulate the scaling function, let a vector $\textbf{Q}_{rm}$ contains all the data points of  the $r$-th sensor under $m$-th working condition. The normalized vector $\hat{\textbf{Q}}_{rm}$ is calculated as follows: 
\begin{align}
\hat{\textbf{Q}}_{rm}=\frac{\textbf{Q}_{rm}-min(\textbf{Q}_{rm})}{max(\textbf{Q}_{rm})-min(\textbf{Q}_{rm})}.
\end{align}


\subsection{Experimental Settings and Evaluation Metrics}

\subsubsection{Experimental Settings} Our architecture is composed of three main parts, namely, encoder network, decoder network, and RUL predictor network. Both encoder and decoder networks rely on LSTM model. To reconstruct the next input sample, the decoder network is followed by a single layer fully connected (FC) network to map from the hidden dimension to the output dimension. The attention mechanism is implemented by two FC networks, i.e., one network computes the attention weights with dimension of $n\times30$, while the other network generates a weighted sum of the encoder hidden states using attention weights. Finally, the RUL predictor network consists of three FC layers, and each layer is followed by rectified linear unit (ReLU) to increase complexity. Adam optimizer is used to optimize the overall model with learning rate of $3e-4$. Moreover, dropout regularization algorithm is employed to relieve the over-fitting problem. Table ~\ref{tab:hyp} summarizes all the hyper-parameters in our ATS2S model.

 \begin{table}[]
\caption{Hyper-parameters of proposed approach}
\label{tab:hyp}
\center
\resizebox{0.48\textwidth}{!}{
\begin{tabular}{@{}ll@{}}
\toprule
Hyper-parameters & Range \\ \midrule
Batch size & \{10\} \\ \midrule
Learning rate & \{0.0003\} \\ \midrule
Training epochs & \{10, 20\} \\ \midrule
Dropout rate & \{0.2, 0.5\} \\ \midrule
Sequence length & \{30\} \\ \midrule
Number of layers (Encoder and Decoder) & \{1\} \\ \midrule
Number of hidden units (Encoder and Decoder) & \{18, 32\} \\ \midrule
Number of layers (Attention Model) & \{2\} \\ \midrule
Number of hidden units (Attention Model) &  L1\{30\}, L2\{9, 14\} \\ \midrule
Number of layers (RUL predictor) & \{2, 3\} \\ \midrule
Number of hidden units (RUL predictor) & L1\{18,32\}, L2\{18,1\}, L3\{1\} \\ \bottomrule
\end{tabular}}
\end{table}

\subsubsection{Performance Metrics} We employ two standard metrics, namely root mean square error (RMSE) and the Score, to evaluate the performance of various methods for RUL prediction. RMSE is defined in Equation (\ref{equ:rmse}).
\begin{align}
RMSE= \sqrt{\frac{1}{N}\sum_{i=1}^N(\widehat{RUL}_i-RUL_i)},
\label{equ:rmse}
\end{align}
where $\widehat{RUL}_i$ and $RUL_i$ are the predicted RUL and true RUL respectively, and $N$ is the total number of samples. For machine prognosis and RUL prediction, late prediction of RUL (e.g., the predicted RUL is longer than the actual RUL) can lead to catastrophic consequences compared to early prediction. However, RMSE is not able to distinguish between early and late predictions. Thus, the Score defined in Equation (\ref{equ:score}) is utilized to give more penalties for late predictions.

\begin{align}
 Score =\left\{
    \begin{array}{ll}
    \frac{1}{N}\sum_{i=1}^N(e^{\frac{\widehat{RUL}_i-RUL_i}{13}-1}),
    &\; \text{if}\;(\widehat{RUL}_i<RUL_i)\\
    \frac{1}{N}\sum_{i=1}^N(e^{\frac{\widehat{RUL}_i-RUL_i}{10}-1}),
    &\; \text{if} \; (\widehat{RUL}_i>RUL_i)\\
    \end{array}
    \right.
    \label{equ:score}
\end{align}

\subsection{Comparison Against State-of-the-arts}

\begin{table*}[ht!]
\center 
\caption{Comparison among various methods in terms of RMSE and Score}
\label{tab:rmse}
\begin{tabular}{@{}llllll@{}@{}lllllll@{}}

\toprule
\multirow{1}{*}{Category}&&\multicolumn{4}{c}{{RMSE}}&&&\multicolumn{4}{c}{{Score}} \\ 
\cmidrule(lr) {3-6} \cmidrule(lr){9-12}  \midrule

& Method& FD001 & FD002 & FD003 & FD004 &&& FD001 & FD002 & FD003 & FD004 \\
\midrule

\multirow{3}{*}{Traditional ML} & SVM~\cite{babu2016deep,dbn} & 40.72 & 52.99 & 46.32 & 59.96 &&& 7703 & 316483 & 22542 & 141122 \\
&RF~\cite{babu2016deep,dbn} & 17.91 & 29.59 & 20.27 & 31.12&&& 480 & 70457 & 711 & 46568\\ 
&GB~\cite{babu2016deep,dbn} & 15.67 & 29.09 & 16.84 & 29.01 &&& 474 & 87280 & 577 & 17818\\
\midrule

\multirow{2}{*}{CNN methods}&2D CNN~\cite{babu2016deep} & 18.45 & 30.29 & 19.82 & 29.16 &&& 1287 & 13570 & 1596 & 7886 \\ 
&1D CNN~\cite{dcnn} & \textbf{12.61} & 22.36 & 12.64 & 23.31 &&& 274 & 10412 & 284 & 12466  \\ 
\midrule

\multirow{3}{*}{LSTM methods}&D-LSTM~\cite{lstm} & 16.14 & 24.49 & 16.81 & 28.17 &&& 338 & 4450 & 852 & 5550  \\ 
&LSTMBS~\cite{lstmbs} & 14.89 & 26.86 & 15.11 & 27.11 &&& 481 & 7982 & 493 & 5200 \\ 
&BLSTM~\cite{Bilstm} & N/A & 25.11 & N/A & 26.61 &&& N/A & 4793 & N/A & 4971\\ 
\midrule
\multirow{1}{*}{Ensemble methods}&MODBNE~\cite{dbn} & 15.04 & 25.05 & 12.51 & 28.66 &&& 334 & 5585 & 422 & 6558\\ \midrule
\multirow{1}{*}{\RaggedRight Encoder-decoder methods}

&BiLSTM-ED~\cite{bilstm-ed} &14.74 &22.07& 17.48& 23.49  &&& 273 & 3099 & 574& 3202\\ 
\midrule

\multirow{3}{*}{Hybrid CNN-LSTM methods}
&CNN-LSTM~\cite{wu2019a} & 14.4 & 27.23 & 14.32 & 26.69 &&& 290 & 9869 & 316 & 6594 \\
&BLCNN~\cite{liu2019a} & 13.18 & 19.09 & 16.76 & 20.97 &&& 302 & 1558 & 381 & 3859 \\ 
&HDNN~\cite{hdnn} & {13.02} & \underline{15.24} & \underline{12.22} & \underline{18.16} &&& \underline{245} & \underline{1282} & \underline{288} & \underline{1527}\\ \midrule \midrule
\textbf{Proposed}&\textbf{ATS2S} & \underline{12.63} & \textbf{14.65} & \textbf{11.44} & \textbf{16.66}&&& \textbf{243} & \textbf{876} & \textbf{263} & \textbf{1074} \\ \midrule \midrule 
\textbf{IMP} &  &  & \textbf{3.87\%} & \textbf{6.4\%} & \textbf{8.3\% }&&& \textbf{0.82\%} & \textbf{31.6\%} & \textbf{8.7\%} & \textbf{29.7\% }\\ \bottomrule \bottomrule
\end{tabular}
\label{SOTA}
\end{table*}

In this section, to comprehensively evaluate our proposed ATS2S method, we compare against 14 
state-of-the-art methods, which can be classified into 6 categories as follows.

\begin{itemize}
\item Traditional machine learning (ML) methods. Three shallow models are employed in the comparison, including support vector machine (SVM)~\cite{dbn}, random forest (RF)~\cite{dbn}, and gradient boosting (GB)~\cite{dbn}.
\item CNN based methods. A 2D CNN network was used in ~\cite{babu2016deep} to predict the RUL for turbofan engines, while Li \textit{et. al,} used 1D CNN with multiple channels for RUL prediction \cite{dcnn}. 
\item LSTM based methods. A standard LSTM network \cite{lstm} and a bi-directional LSTM \cite{Bilstm} were developed for RUL prediction. In \cite{lstmbs}, LSTM is augmented with a bootstrap algorithm to predict the RUL values. 
\item Ensemble methods. A deep belief network (DBN) is used together with ensemble techniques for the RUL prediction task~\cite{dbn}.
\item Hybrid CNN-LSTM based methods. Combination of CNN and LSTM models has been used for RUL prediction. CNN and LSTM can be cascaded in a sequential manner, e.g., CNN-LSTM~\cite{wu2019a} put CNN in the first stage, while BLCNN~\cite{liu2019a} reversed the order. In addition, HDNN~\cite{hdnn} combined both the features from CNN and LSTM to generate the final predictions. 
\item Encoder-decoder based methods. BiLSTM-ED \cite{bilstm-ed} first extracts health index and then estimates the health index curves using linear regression model. Finally it uses curve-similarity matching to estimate the RUL.
\end{itemize}

Table~\ref{SOTA} shows the comparison among the above methods for RUL prediction. Note that the highest score in each column is in \textbf{bold}, while the second best score is \underline{underlined}. 

We can observe that our proposed ATS2S outperforms all the other methods consistently, except that it achieves a comparable RMSE with 1D CNN \cite{dcnn} on FD001 dataset. In particular, our ATS2S achieves significant improvement over the state-of-the-arts on FD002 and FD004, which are two complex datasets with multiple working conditions and thus indicate more practical scenarios. For example, ATS2S is able to achieve improvements over the second best performer on FD004 by 8.3\% and 29.7\% in terms of RMSE and Score, respectively. Such improvements on FD002 and FD004 demonstrate that ATS2S has clear advantages over the competing methods to handle the complex datasets. In addition, compared with the RMSE metric, our ATS2S achieves even better improvements in terms of the Score metric, indicating that we can better address the issue of late predictions.

\subsection{Model Analysis}

\subsubsection{Ablation Study} 
\begin{figure}[htbp]
\centering
\begin{minipage}{0.24\textwidth}
    \centering
    \includegraphics[width=0.98\textwidth,height=90mm,keepaspectratio]{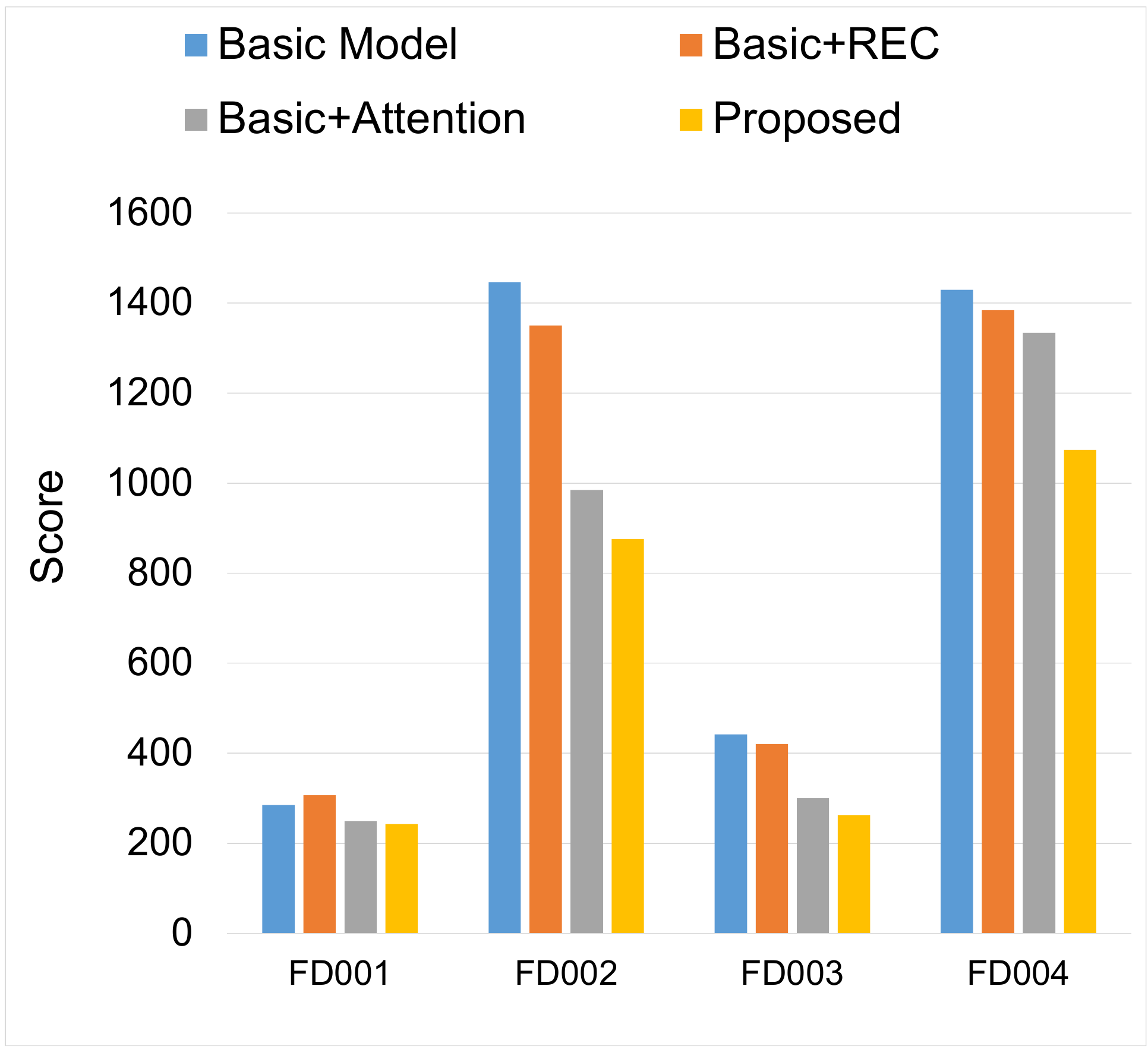}
    \subcaption{}
\end{minipage}%
\begin{minipage}{0.24\textwidth}
    \centering
    \includegraphics[width=0.98\textwidth,height=90mm,keepaspectratio]{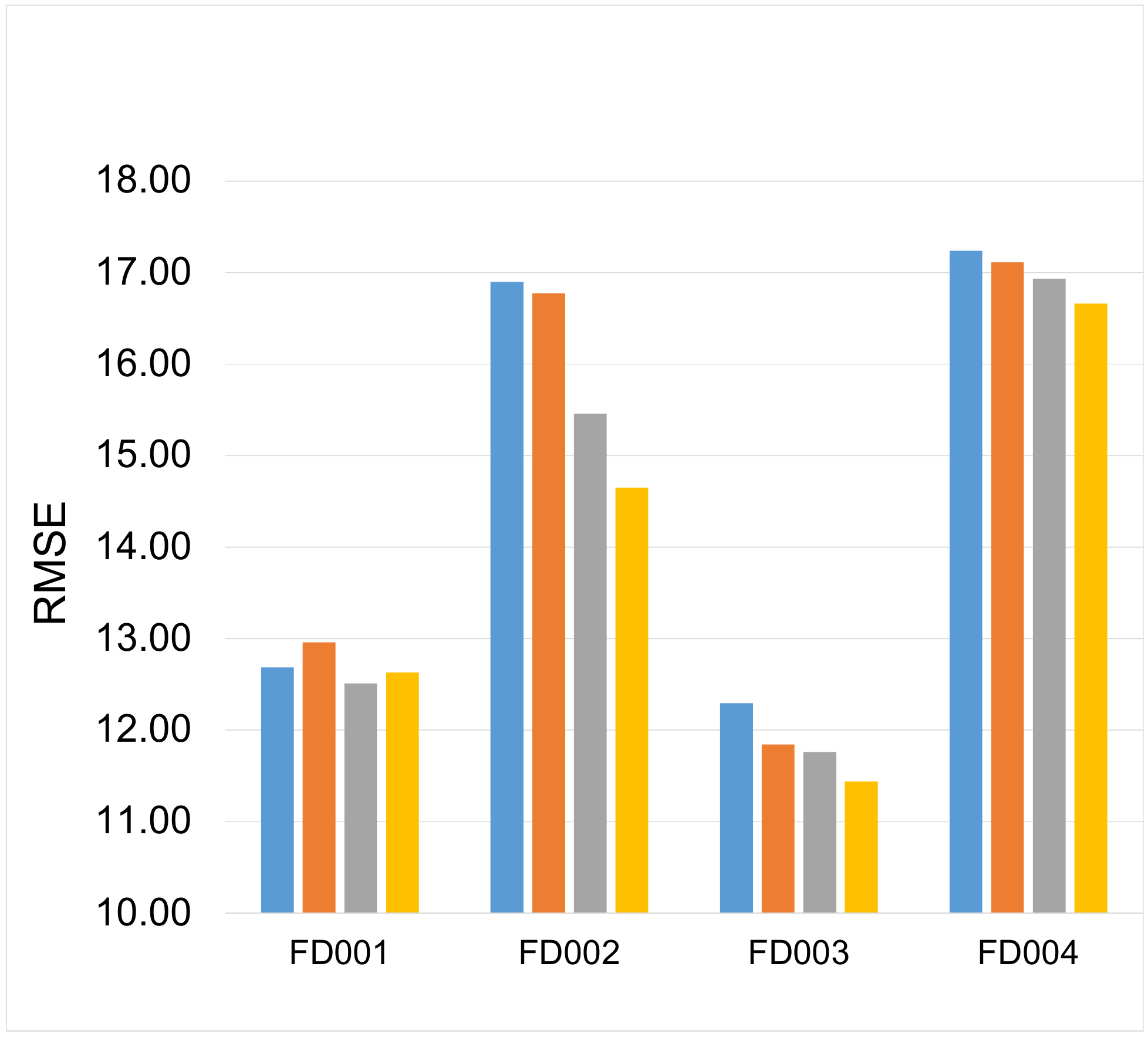}
    \subcaption{}
\end{minipage}%
\caption{Ablation study for the proposed ATS2S method} \label{fig:ablation}
\end{figure}

In this section, we disentangle the contribution of each part of the ATS2S model. In addition to the ATS2S model, we further derive three variants, namely (1) Basic sequence to sequence model without reconstruction or attention, (2) Basic model with reconstruction, (3) Basic model with attention.  Fig.~\ref{fig:ablation} shows the comparison between these 3 variants and the proposed ATS2S model. Based on the comparison shown in Fig.~\ref{fig:ablation}, we can further draw two conclusions.

Firstly,  our proposed ATS2S model with both attention mechanism and reconstruction architecture achieves the best performance over 4 datasets in terms of both metrics, showing that it is indeed more effective for RUL prediction than basic sequence to sequence model. 
This demonstrates that learning from most relevant information from long sensor signals by attention mechanism (not just focusing on the latest information) , as well as enabling predictive power and capturing temporal dependencies by reconstruction architecture, are critical for improving RUL prediction. 

Secondly, the model with attention mechanism outperforms the model with reconstruction architecture, indicating that attention mechanism has larger impact than reconstruction task in our ATS2S model. Without the attention mechanism, we squash the whole input sequence into a single hidden vector (i.e., the \textit{last} hidden state of the encoder). Instead, attention mechanism can consider all the hidden states with different weights and help to learn better comprehensive \textit{dual-latent feature representation} from both encoder and decoder for RUL prediction.

\begin{figure}[!hb]
\centering
\begin{minipage}{0.23\textwidth}
  \centering
\includegraphics[width=0.95\textwidth,height=80mm,keepaspectratio]{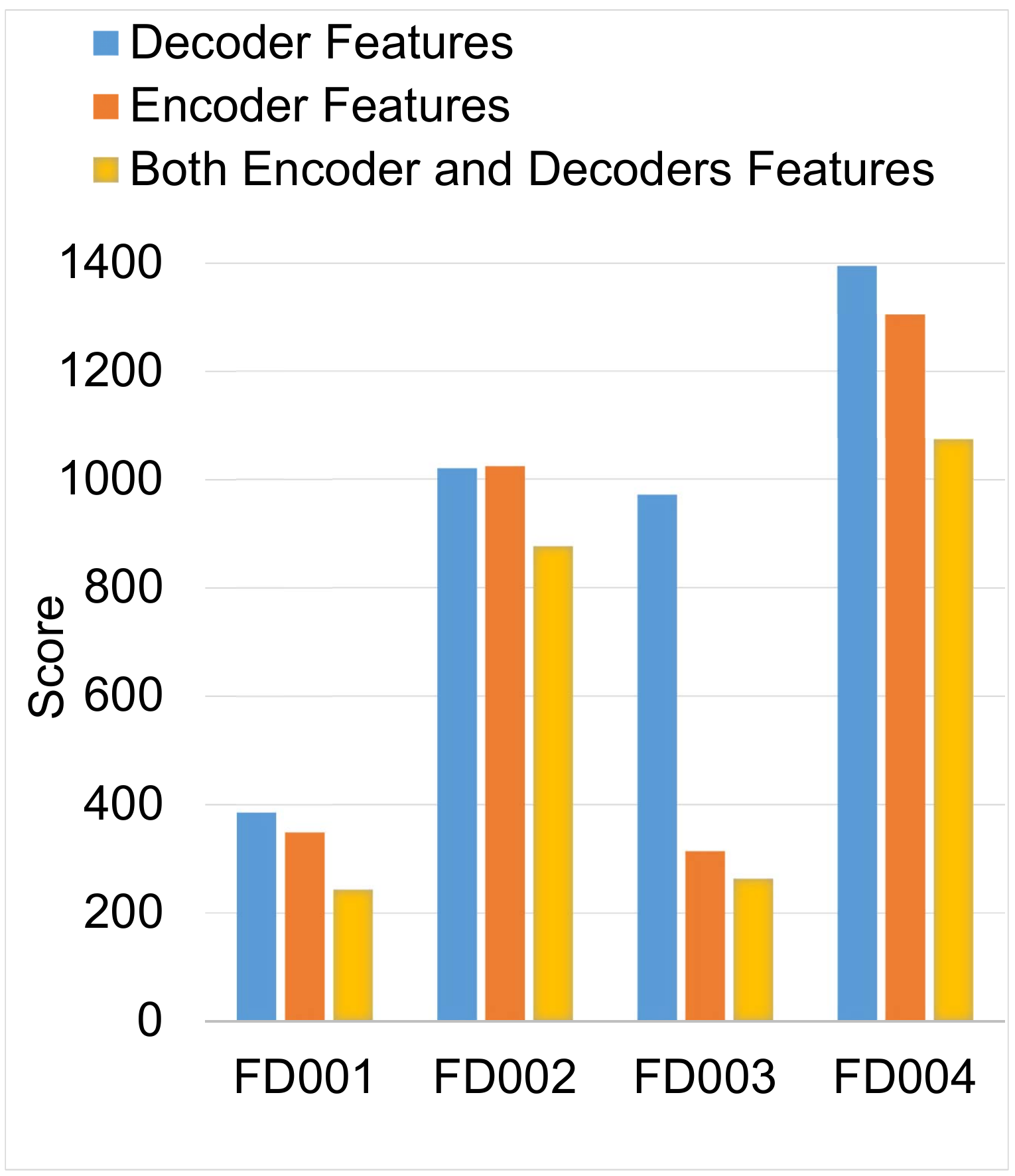}
\subcaption{}
\end{minipage}%
\begin{minipage}{0.23\textwidth}
  \centering
\includegraphics[width=0.95\textwidth,height=80mm,keepaspectratio]{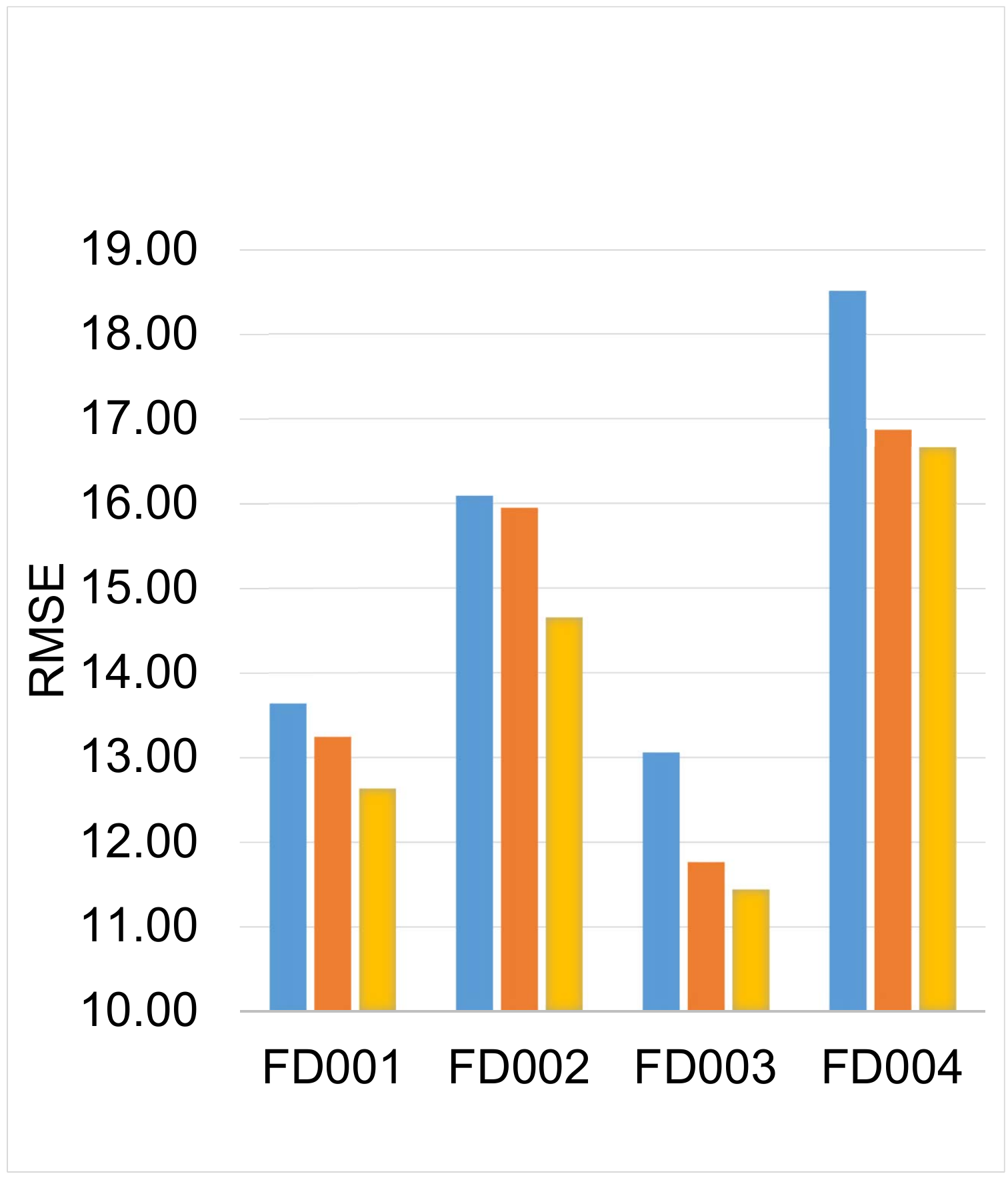}
\subcaption{}
\end{minipage}%
\caption{Study of feature importance of the proposed method} \label{fig:fe}
\end{figure}

\subsubsection{Feature Importance Analysis} As shown in Fig. \ref{rul_pred}, we use the \textit{dual-latent feature representation} to integrate features from both encoder and decoder for RUL prediction. To study the importance of the features used in our ATS2S, we conduct experiments using three different feature sets, namely, encoder features (i.e., encoder hidden states), decoder features and integrated features, i.e., encoder-decoder features (dual-latent feature representation). Fig.~\ref{fig:fe} shows the detailed model performance with three different feature sets. We can observe that dual-latent feature representation achieves the best performance over all four data subsets consistently, indicating the importance of a comprehensive representation with rich semantics from both encoder and decoder features.

\subsubsection{Sensitivity Analysis} 
\begin{figure}[!h]
\centering
\begin{minipage}{0.24\textwidth}
  \centering
\includegraphics[width=0.98\textwidth,height=40mm]{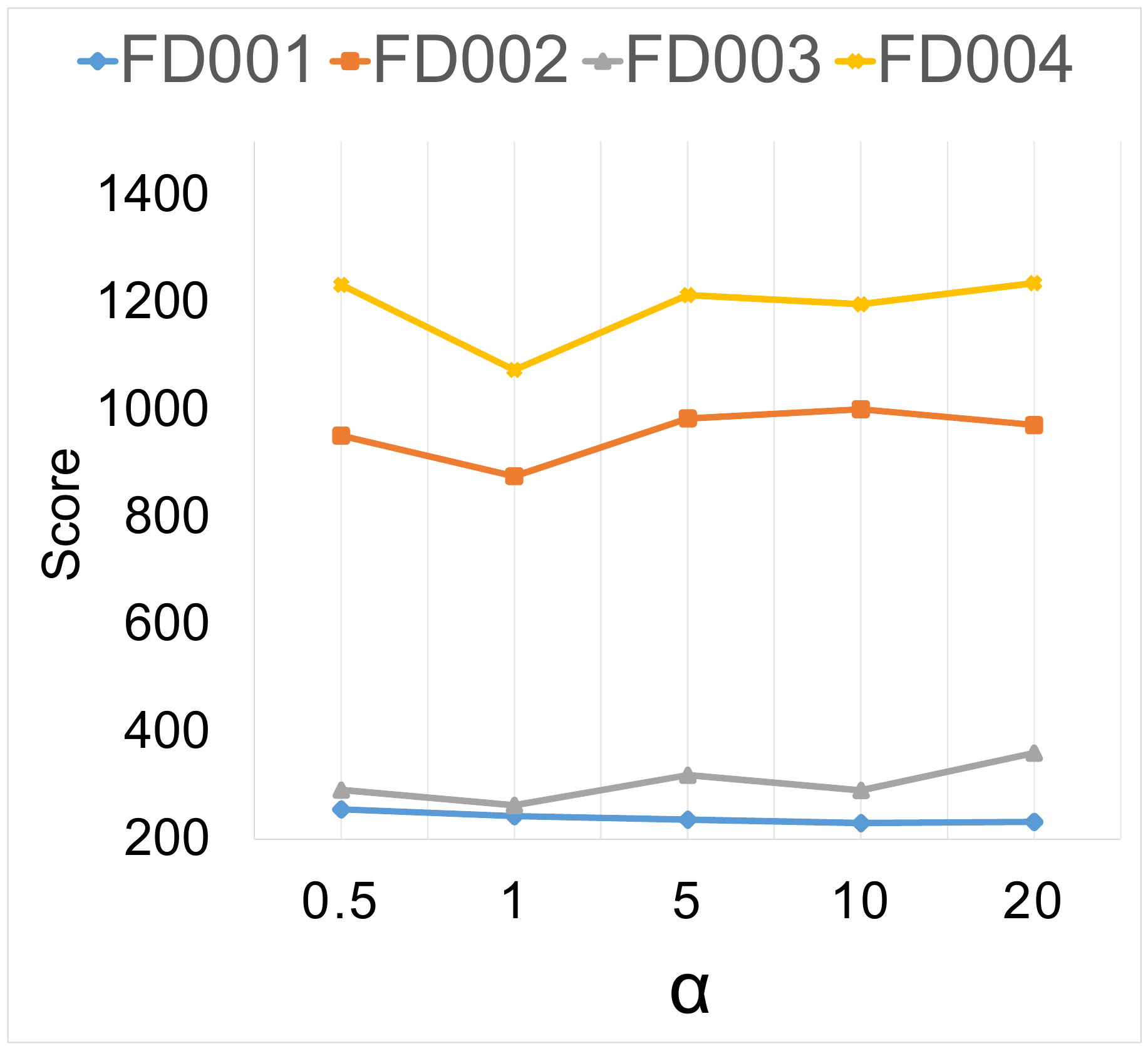}
\subcaption{}
\end{minipage}%
\begin{minipage}{0.24\textwidth}
  \centering
\includegraphics[width=0.98\textwidth,height=40mm]{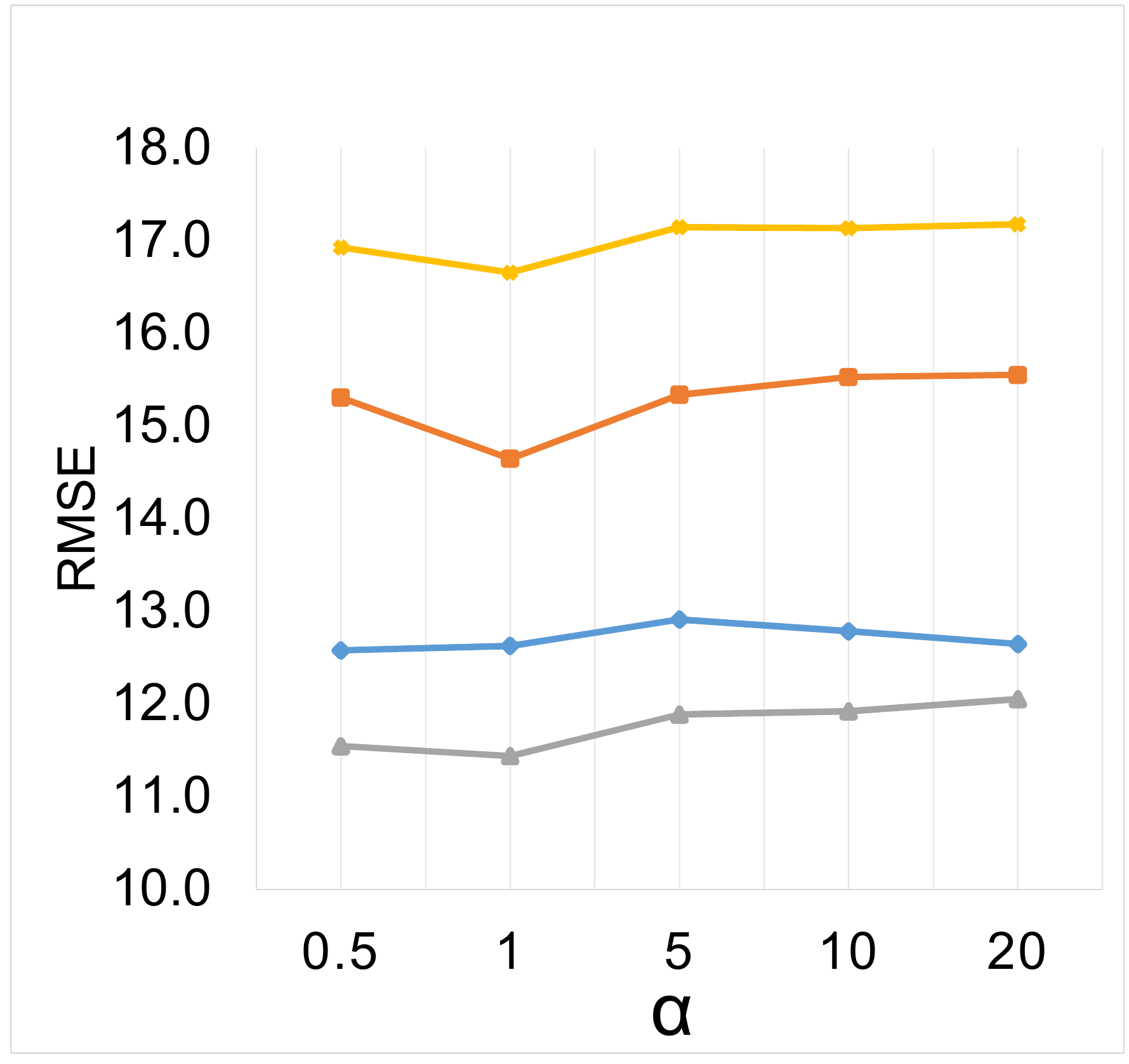}
\subcaption{}
\end{minipage}%
\caption{Sensitivity analysis of reconstruction weight} \label{fig:se}
\end{figure}

As shown in Equation (\ref{joint_loss}), the parameter $\alpha$ controls the contribution of reconstruction loss in the final joint loss. In this section, we perform the sensitivity analysis for this parameter $\alpha$. Fig.~\ref{fig:se} shows the performance of ATS2S model across four datasets with different values for $\alpha$. Overall, it can be clearly observed that equal contribution from both reconstruction and prediction loss (i.e., $\alpha =1$) achieves the best performance, demonstrating that both of them are critical for accurate RUL predictions.

\section{Conclusion}
In this work, we presented a novel attention-based sequence to sequence model ATS2S to accurately predict equipment RUL, which has huge impact for many real-world applications.  
In particular, we designed a novel framework that learns to reconstruct the next sequence and predict the RUL labels concurrently. In addition, we showed our attention mechanism can better capture all the relevant historical information from long sensor sequences than standard LSTM approach which focuses on the latest information only. Finally, our   \textit{dual-latent feature representation} which integrate both the encoder and decoder features is very effective for RUL prediction. Our extensive experimental results demonstrate that our proposed ATS2S significantly outperforms 13 state-of-the-arts for RUL prediction across 4 benchmark datasets consistently.

\bibliographystyle{Bibliography/IEEEtran}
\bibliography{ATS2S}

\end{document}